\documentclass[runningheads]{llncs}

\usepackage{graphicx}

\usepackage{amssymb}
\usepackage{amsmath}

\DeclareMathOperator*{\argmax}{argmax}

\begin{document}
\title{Posterior Adaptation With New Priors} 
\titlerunning{Posterior Adaptation With New Priors}

\author{Jim Davis}
\authorrunning{J. Davis}

\institute{Dept. Computer Science and Engineering\\
Ohio State University\\
Columbus OH 43210 USA\\
\email{davis.1719@osu.edu}}

%
%
\maketitle              
\begin{abstract} 
Classification approaches based on the direct estimation and analysis of posterior probabilities will degrade over time if the original class priors begin to change. We prove that a unique (up to scale) solution is possible to recover the data likelihoods for a test example from its estimated class posteriors and the original dataset priors. Given the recovered likelihoods and a set of new priors, the posteriors can be re-computed using Bayes' Rule to reflect the influence of the new priors. The method is simple to compute and allows a dynamic update of the original posterior model.
\end{abstract}
%
%
%

\section{Introduction}

A common probabilistic approach to classification in Machine Learning is the {\em Maximum A Posteriori} (MAP) technique \cite{PatternRecognitionAndMachineLearning} in which the class $k$ having the largest posterior probability with test example $T$ is selected
\begin{equation}
k_{MAP} = \argmax_i P(C_i | T)
\end{equation}

\noindent One approach to MAP is to estimate the data likelihood for each class $P(T | C_i)$ and use the class priors $P(C_i)$ within Bayes' Rule to compute the posterior
\begin{equation}
P(C_i | T)  =  \frac{P(T | C_i)P(C_i)}{\sum_j P(T | C_j)P(C_j)} 
\label{eqn:bayes-rule}
\end{equation}

\noindent If only the resulting maximal class is desired, and not the actual posterior itself, the denominator in Eqn.~\ref{eqn:bayes-rule} need not be calculated as it is constant across all classes. However, in some situations it is not clear how to model or learn an appropriate likelihood model.

Alternatively, one can compute the posterior directly, without specific knowledge or application of the likelihoods or priors. 
One simple approach with sufficient training data is based on K-nearest neighbors, where the proportions of each class within the set of K-nearest training examples to a test example can be used to locally estimate class posteriors \cite{StatisticalPatternRecognition}. In \cite{Hedging}, several one-vs-all SVM classifiers are initially trained and the output from each SVM is modeled as a posterior using Platt Scaling (logistic regression with a single logit). The set of posteriors are L1-normalized to get multi-class posterior estimates for the test example. In \cite{HierarchicalLogits}, a posterior method is outlined with logistic regression on the full set of neural network output logits corresponding to the target classes. As the output of modern neural networks is often uncalibrated (i.e., the  softmax value for the argmax class does not properly represent the accuracy of classification)  \cite{NetworkCalibration}, any post-calibration of the argmax-selected output can also be considered a posterior model. In these (and other) approaches, the posteriors are computed without explicit use of any likelihood or prior components in the model (relying on the data distribution itself during training).

One issue with these ``direct'' approaches when deployed in practice is that the posterior model can give inaccurate results if the class priors change over time due to external factors. To aid in the flexibility of direct posterior-based approaches within a dynamic domain, we present a technique to uniquely recover (up to scale) the corresponding class likelihoods of a test example given its  class posteriors from the initial model and also the original data priors. The priors can simply be represented by the class frequencies in the original dataset and no specific proportion relationship is required (i.e., the priors need not be equal).
The recovered likelihoods and a set of new priors can then be used within Bayes' Rule (Eqn.~\ref{eqn:bayes-rule}) to compute the updated posteriors to be reflective of the new situation.  


\section{Approach}

Assume a classifier is trained on a dataset with the following class priors and posteriors for test example $T$
\begin{eqnarray}
P(C_i)>0 &, & \sum_i P(C_i) =1\\
P(C_i | T)>0 &, & \sum_i P(C_i | T) =1
\end{eqnarray}
\noindent Bayes' Rule (Eqn.~\ref{eqn:bayes-rule}) on class $C_i$ gives the posterior as a combination of likelihoods $P(T | C_j)>0$ and given priors. Manipulation of the terms in Eqn.~\ref{eqn:bayes-rule} yields
\begin{eqnarray}
P(C_i | T)\sum_jP(T | C_j)P(C_j) - P(T | C_i)P(C_i) & = & 0\\
(P(C_i | T)-1)P(T | C_i)P(C_i)+\sum_{j \neq i} P(C_i | T)P(T | C_j)P(C_j) & = & 0
\end{eqnarray}

Doing this for each class posterior and factoring out the likelihoods yields a homogeneous set of linear equations $M{\vec x} = {\vec 0}$ with
\begin{equation}
M = 
\begin{bmatrix}
(P(C_1 | T)-1)P(C_1) & P(C_1 | T)P(C_2) & \cdots & P(C_1 | T)P(C_n)\\
P(C_2 | T)P(C_1) & (P(C_2 | T)-1)P(C_2) & \cdots & P(C_2 | T)P(C_n)\\
 \vdots & \vdots & \vdots & \vdots \\
P(C_n | T)P(C_1) & P(C_n | T)P(C_2) & \cdots & (P(C_n | T)-1)P(C_n)
\end{bmatrix}
\end{equation}

\begin{equation}
{\vec x} = 
\begin{bmatrix}
P(T | C_1)\\
P(T | C_2)\\
\vdots\\
P(T | C_n)
\end{bmatrix}
\end{equation}

 Therefore the likelihoods are the non-trivial (${\vec x} \neq {\vec 0}$) solution to this homogeneous linear system. However, {\it does an exact positive solution to ${\vec x}$ exist}? We will show that the answer is yes, and thus enable computation of updated posteriors given new/evolving class priors. Note that the solution to ${\vec x}$ in $M{\vec x} = 0$ will be valid up to a scalar, as $M(\alpha{\vec x}) = 0$, however this will not be of concern when updating the posteriors due to the normalization within Bayes' Rule.

\subsection{Proof of a positive, exact solution}

Inspection of matrix $M$ shows that it has positive off-diagonal elements, 
but negative diagonal elements bound by $-1 < M_{i,i} < 0$. 
Consider matrix $A = M + I$ 

\begin{equation}
\small
A = 
\begin{bmatrix}
(P(C_1 | T)-1)P(C_1)+1 & P(C_1 | T)P(C_2) & \cdots & P(C_1 | T)P(C_n)\\
P(C_2 | T)P(C_1) & (P(C_2 | T)-1)P(C_2)+1 & \cdots & P(C_2 | T)P(C_n)\\
 \vdots & \vdots & \vdots & \vdots \\
P(C_n | T)P(C_1) & P(C_n | T)P(C_2) & \cdots & (P(C_n | T)-1)P(C_n)+1
\end{bmatrix}
\end{equation}

\noindent where $A$ is a real positive (all $A_{i,j} > 0$) square matrix.

Perron-Frobenious Theory \cite{PerronFrobenious} states that any positive real square matrix has a single {\em positive} eigenvector that corresponds to its {\em largest} eigenvalue (which is positive and real). Furthermore, any other eigenvalue (possibly complex) in absolute value is strictly smaller than the maximum eigenvalue. Therefore $A$ has a positive eigenvector which will be shown to be the solution to ${\vec x}$ (the desired likelihoods, up to scale).

Following the theory applied to $A$, there is a positive eigenvector $v^A_+$ corresponding to the maximum eigenvalue $\lambda^A_{max}$ such that
\begin{equation}
A{\vec v^A_{+}} = \lambda^A_{max} \cdot {\vec v^A_{+}} \label {eqn:eigen_eqn}
\end{equation}

\noindent The theorem further states that the maximum eigenvalue $\lambda^A_{max}$ is bound by the minimum and maximum row sums of $A$:
\begin{equation}
\min_i \sum_j A_{i,j} \leq \lambda^A_{max} \leq \max_i \sum_j A_{i,j}
\end{equation}

Since, the eigenvalues of a matrix and its transpose are the same, the maximum eigenvalue of $A$ is the maximum eigenvalue of $A^T$ (i.e., $\lambda^A_{max} = \lambda^{A^T}_{max}$). Therefore, we can equivalently use the minimum and maximum {\em column} sums 
\begin{equation}
\min_j \sum_i A_{i,j} \leq \lambda^A_{max} \leq \max_j \sum_i A_{i,j}
\end{equation}

\noindent as it is trivial to show that $\sum_i A_{i,j} = 1$ for any column $j$
\begin{eqnarray}
\sum_i A_{i,j} & = & (P(C_j|T) -1)P(C_j) + 1 + \sum_{i \neq j} P(C_i | T)P(C_j) \\
 & = & 1 - P(C_j) + \sum_{i} P(C_i | T)P(C_j)\\
 & = & 1 - P(C_j) + P(C_j) \\
& = & 1
\end{eqnarray}

\noindent Thus a single maximum eigenvalue $\lambda^A_{max} = 1$ exists for $A$.

Substituting $M + I = A$ in Eqn. \ref{eqn:eigen_eqn} gives
\begin{eqnarray}
\left(M+I\right){\vec v^A_{+}} & = & \lambda^A_{max} \cdot {\vec v^A_{+}}\\
M{\vec v^A_{+}} & = & \lambda^A_{max} \cdot {\vec v^A_{+}} - {\vec v^A_{+}}\\
M{\vec v^A_{+}} & = & \left(\lambda^A_{max}-1\right) \cdot {\vec v^A_{+}}\\
M{\vec v^A_{+}} & = & \rho \cdot {\vec v^A_{+}}
\end{eqnarray}
\noindent where $\rho = \left(\lambda^A_{max}-1\right)$. Since $\lambda^A_{max}=1$, the value of $\rho = \left(\lambda^A_{max}-1\right) = 0$. Therefore $\rho=0$  provides the optimized solution to $M{\vec x}=0$ for a non-trivial vector ${\vec x} = {\vec v^A_{+}}$.  

This proof shows the existence and guarantee of a positive vector ${\vec v^A_{+}}$ (the likelihoods, up to scale) with zero error ($\rho = 0$). The computed eigenvector is typically normalized to a unit vector, but as mentioned, its use within Bayes' Rule will cancel any scaling. 


\subsection{Relationship to Non-Square Matrix Solution}

Given $B{\vec x} = 0$ with non-square matrix $B$, the classic approach (as commonly used in camera calibration techniques \cite{Hartley}) to solving the non-trival solution to ${\vec x}$  is to find the eigenvector corresponding to the {\em smallest} eigenvalue of $B^TB$ in
\begin{equation}
\left(B^TB\right){\vec x} = {\vec 0}
\end{equation}

\noindent This form is derived from a Lagrangian loss function based on minimizing $||Bx||^2 = \left(B{\vec x}\right)^T(B{\vec x}) = {\vec x}^TB^TB{\vec x}$ subject to $||{\vec x}||^2 = {\vec x}^T{\vec x} = 1$ \cite{Hartley}. 

Though our matrix $M$ is square, we can easily confirm the above non-square formulation holds by substituting $M = B$ and ${\vec v^A_+} = {\vec x}$ into the minimization loss (we already know that $||{\vec v^A_+}||^2 = 1$)
\begin{eqnarray}
 ||M {\vec v^A_+}||^2  & = & (M {\vec v^A_+})^T(M {\vec v^A_+}) \\
                        & = & (\rho {\vec v^A_+})^T(\rho {\vec v^A_+})\\
                        & = & \rho^2 ({\vec v^A_+})^T {\vec v^A_+}\\
                        & = & \rho^2 
\end{eqnarray}

\noindent As $\rho^2=0$, the loss is minimal. Thus the required positive eigenvector can be computed directly from matrix $A$ without the need for any matrix transpose multiplication.


\subsection{Posterior Updating Procedure}

To update the existing posterior estimation for a test example with new priors, we form matrix $A$ using the current posterior estimates $P(C_i | T)$ and the original dataset priors $P(C_i)$. Then the  eigenvector corresponding to the maximum eigenvalue of $A$ is computed, which is the vector of  scaled likelihoods $P(T | C_i)$.  Bayes' Rule is applied with these likelihoods and
a set of new priors ${\hat P}(C_i)$ to provide the updated posteriors ${\hat P}(C_i | T)$ for $T$
\begin{equation}
{\hat P}(C_i | T)  =  \frac{P(T | C_i){\hat P}(C_i)}{\sum_j P(T | C_j){\hat P}(C_j)} 
\end{equation}

\noindent These updated posteriors reflect the new class priors and can be used in any further decision making process.


\section{Example}

\begin{figure}[!t]
\centering
\setlength{\tabcolsep}{0.0pt}
\begin{tabular}{cc}
\includegraphics[height=1.7in]{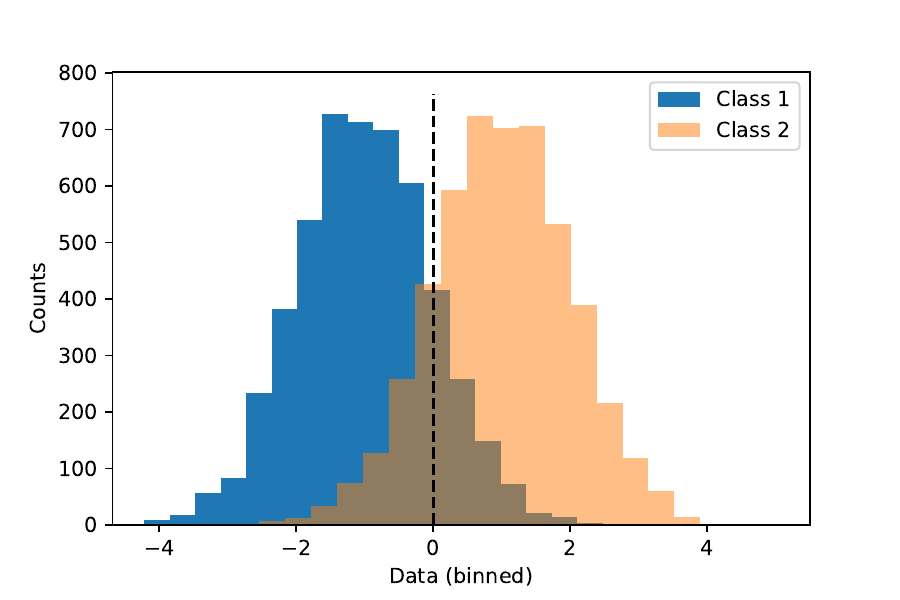} & 
\includegraphics[height=1.7in]{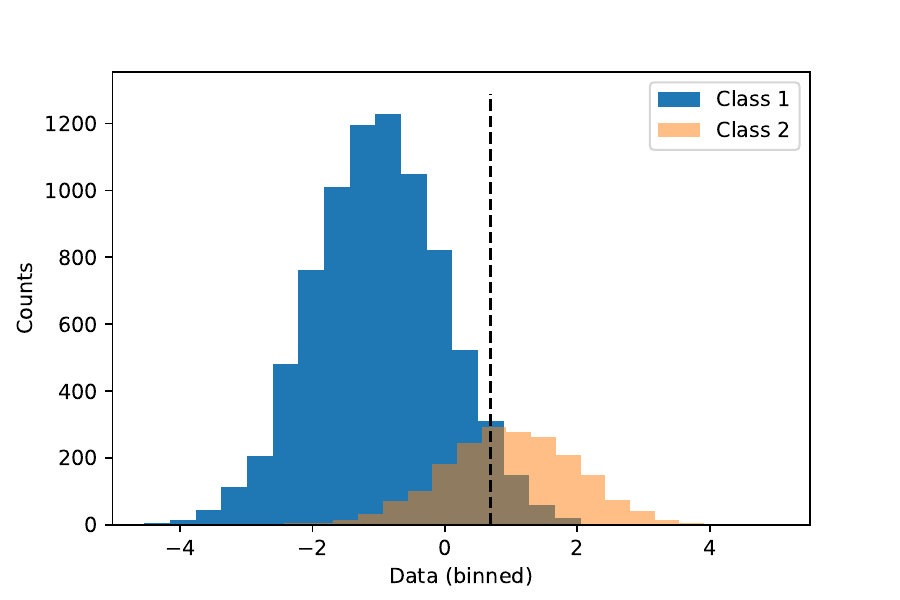} \\
(a) & (b) 
\end{tabular}
\caption{Data distributions with (a) original priors and (b) new priors.}
\label{fig:hists}
\end{figure}

To demonstrate the method, we create a set of synthetic data from which the known posteriors, likelihoods, and priors can be used and verified. We will show how the given posteriors for a test example can be easily modified with new priors using the proposed method.

We initially create a 1-D Gaussian likelihood distribution for each of two classes ($C_1$, $C_2$) using
\begin{equation}
P(d | \mu_{C_i}, \sigma) = \frac{1}{\sqrt{2\pi \sigma^2}} \exp{-\frac{(d-\mu_{C_i})^2}{2\sigma^2}}
\end{equation}

\noindent with $\mu_{C_1} = -1.0$, $\mu_{C_2} = 1.0$, and $\sigma^2  = 1.0$. In this example we begin with equal priors $P(C_1) = P(C_2) = .5$. Sampling a total of $N=10K$ values provides the dataset having the histogram distributions shown in Fig.~\ref{fig:hists}(a). 

We simulate a Bayesian classifier that directly computes the posterior probabilities for a test sample and then chooses the class with the largest posterior. We use the optimal posteriors for each sample by using Bayes Rule (Eqn. \ref{eqn:bayes-rule}) with the ground truth Gaussian likelihood models and priors (though the likelihoods would not be available to a direct, posterior-based classifier). Given the Gaussian likelihood distributions and the priors, the optimal Bayesian classification boundary for our class means and unit variance is the location $d$ where the two probabilities are equal
\begin{eqnarray}
P(C_1 | d) & = & P(C_2 | d)\\
P(d | C_1)P(C_1) & = & P(d | C_2)P(C_2)\\
\frac{P(d | C_1)}{ P(d | C_2)} & = &\frac{P(C_2)}{P(C_1)}\\
\ln{\frac{P(d | C_1)}{ P(d | C_2)} }& = & \ln{\frac{P(C_2)}{P(C_1)}}\\
\frac{-(d-\mu_{C_1})^2}{2\sigma^2} - \frac{-(d-\mu_{C_2})^2}{2\sigma^2} & = & \ln{\frac{P(C_2)}{P(C_1)}}\\
(d+1)^2 - (d-1)^2 & = & -2 \ln{\frac{P(C_2)}{P(C_1)}}\\
d & = & -\frac{1}{2}\ln{\frac{P(C_2)}{P(C_1)}}
\end{eqnarray}

\noindent This boundary for equal priors is $d=0$ and is shown as the vertical dashed line in Fig.~\ref{fig:hists}(a).

Let us assume that some time in the future the class priors have changed, e.g.,  $\hat{P}_{C_1} = .8$ and $\hat{P}_{C_2} = .2$, and we collect $N$ new data samples. The corresponding histogram distributions and optimal Bayesian decision threshold for this new collection are shown in Fig.~\ref{fig:hists}(b). Notice that the new optimal boundary ($\hat{d} = .69$) has shifted to the right due to the stronger prior of $\hat{P}_{C_1}$. If we use the maximum of the posteriors computed from our previous posterior model (based on the old priors and corresponding to $d=0$), the total error rate is 15.9\% (1269/8000 errors in $C_1$ and 319/2000 errors in $C_2$). The new data obviously does not reflect the previous equal priors.

Instead, we compute for each new data sample its original class posteriors (based on the previous priors), use the proposed method to recover the likelihoods using these posteriors and previous priors, and then update the posteriors. Classification using the maximum of these new posteriors yields a reduced total error rate of 11.2\% (362/8000 errors in $C_1$ and 756/2000 errors in $C_2$). These are the same results if the optimal Bayesian threshold $\hat{d} = .69$ was used directly on the sample data. Therefore, the proposed approach can be used to properly update the posteriors to achieve more prior-aware classifications.


\section{Summary}

We presented an approach to recover (up to scale) the underlying likelihoods of a test example from its class posteriors and priors. As fixed models become inaccurate over time if the priors begin to change, the method enables an update of the posterior estimates with a set of new class priors.  This approach is applicable to those classification models that directly estimate the posterior of each class, without explicitly using the likelihood or priors (relying instead on the data distribution itself during training). We expect this technique to be useful in variety of dynamic classification scenarios.



%

 \bibliographystyle{splncs04}
 \bibliography{mybib}

\begin{thebibliography}{1}
\providecommand{\url}[1]{\texttt{#1}}
\providecommand{\urlprefix}{URL }
\providecommand{\doi}[1]{https://doi.org/#1}

\bibitem{PatternRecognitionAndMachineLearning}
Bishop, C.: Pattern Recognition and Machine Learning. Springer, New York (2006)

\bibitem{HierarchicalLogits}
Davis, J., Liang, T., Enouen, J., Ilin, R.: Hierarchical classification with
  confidence using generalized logits. In: ICPR (2021)

\bibitem{Hedging}
Deng, J., Krause, J., Berg, A.C., Fei-Fei, L.: Hedging your bets: {O}ptimizing
  accuracy-specificity trade-offs in large scale visual recognition. In: CVPR
  (2012)

\bibitem{StatisticalPatternRecognition}
Fukunaga, K.: Introduction to Statistical Pattern Recognition. Academic Press,
  New York, 2 edn. (1990)

\bibitem{NetworkCalibration}
Guo, C., Pleiss, G., Sun, Y., Weinberger, K.: On calibration of neural
  networks. In: ICML (2017)

\bibitem{Hartley}
Hartley, R.: In defense of the eight-point algorithm. IEEE Trans. Pattern
  Analysis and Machine Intelligence  \textbf{19}(6),  580--593 (1997)

\bibitem{PerronFrobenious}
Meyer, C.: Matrix Analysis and Applied Linear Algebra. SIAM, Philadelphia, PA
  (2000)

\end{thebibliography}

\end{document}